# A High Speed Multi-label Classifier based on Extreme Learning Machines

Meng Joo Er, Rajasekar Venkatesan and Ning Wang

**Abstract.** In this paper a high speed neural network classifier based on extreme learning machines for multi-label classification problem is proposed and discussed. Multi-label classification is a superset of traditional binary and multi-class classification problems. The proposed work extends the extreme learning machine technique to adapt to the multi-label problems. As opposed to the single-label problem, both the number of labels the sample belongs to, and each of those target labels are to be identified for multi-label classification resulting in increased complexity. The proposed high speed multi-label classifier is applied to six benchmark datasets comprising of different application areas such as multimedia, text and biology. The training time and testing time of the classifier are compared with those of the state-of-the-arts methods. Experimental studies show that for all the six datasets, our proposed technique have faster execution speed and better performance, thereby outperforming all the existing multi-label classification methods.

## 1 Introduction

In recent years, the problem of multi-label classification is gaining much importance motivated by increasing application areas such as text categorization [1-5], marketing, music categorization, emotion, genomics, medical diagnosis [6], image and video categorization, etc. Recent realization of the omnipresence of multi-label prediction tasks in real world problems has drawn increased research attention [7].

Classification in machine learning is defined as "Given a set of training examples composed of pairs $\{x_i,y_i\}$, find a function f(x) that maps each attribute vector $x_i$ to its associated class $y_i$, i = 1,2,….,n, where n is the total number of training samples" [8]. These classification problems are called single-label classification. Single-label classification problems involve mapping each of the input vectors to its unique target class from a pool of target classes. However, there are several classification problems in which the target classes are not mutually exclusive and the input samples belong to more than one target class. These problems cannot be classified using single-label classification thus resulting in the development of several multi-label classifiers to mitigate this limitation. By the recent advancements in technology, the application areas of multi-label classifiers spread across various domains such as text categorization, bioinformatics [9-10], medical diagnosis, scene classification [11-12], map labeling [13],

Meng Joo Er is a Chair Professor with Marine Engineering College, Dalian Maritime University, Dalian 116026, China, and together with Rajasekar Venkatesan are with the School of Electrical and Electronics Engineering in NTU, Singapore; Ning Wang is with Marine Engineering College, Dalian Maritime University, Dalian 116026, China.

multimedia, biology, music categorization, genomics, emotion, image and video categorization and so on. Several classifiers are developed to address the multi-label problem and are available in the literature. Multi-label problems are more difficult and more complex compared to single-label problems due to its generality [14]. In this paper, we propose a high-speed multi-label classifier based on extreme learning machines (ELM). The proposed ELM-based approach outperforms all existing multi-label classifiers with respect to training time and testing time and other performance metrics.

The rest of the paper is organized as follows. A brief overview of different types of multi-label classifiers available in the literature is discussed in Section II. Section III describes the proposed approach for multi-label problems. Different benchmark metrics for multi-label datasets and experimentation specifications are discussed in Section IV. In Section V, a comparative study of the proposed method with existing methods and related discussions are carried out. Finally, concluding remarks are given in Section VI.

## 2 Multi-label Classifier

The definition for multi-label learning as given by [15] is; "Given a training set, S = ($x_i$, $y_i$), $1 \le i \le n$, consisting of n training instances, ($x_i \in X$, $y_i \in Y$) drawn from an unknown distribution D, the goal of multi-label learning is to produce a multi-label classifier $h:X\rightarrow Y$ that optimizes some specific evaluation function or loss function".

Let $p_i$ be the probability that the input sample is assigned to $i^{th}$ class from a pool of M target classes. For single-label classification such as binary and multi-class classification the following equality condition holds true.

$$\sum p_i = 1 \tag{1}$$

This equality does not hold for multi-label problems as each sample may have more than one target class. Also, it can be seen that the binary classification problems, the multi-class problems and ordinal regression problems are specific instances of the multi-label problems with the number of labels corresponding to each data sample restricted to 1 [16].

The multi-label learning problem can be summarized as follows:

- There exists an input space that contains tuples (features or attributes) of size D of different data types such as Boolean, discrete or continuous. $x_i \in X$, $x_i = (x_{i1}, x_{i2}, \ldots x_{iD})$.
- A label space of tuple size M exists which is given as, $L = \{\zeta_1, \zeta_2, \ldots, \zeta_M\}$ .
- Each data sample is given as a pair of tuples (input space and label space respectively). $\{(x_i,y_i) \mid x_i \in X, y_i \in Y, Y \subseteq L, 1 \le i \le N\}$ where N is the number of training samples.
- A training model that maps the input tuple to the output tuple with high speed, high accuracy and less complexity.

Several approaches for solving multi-label problem are available in the literature. Earlier categorization of the multi-label (ML) methods [17] classify the methods into two categories, namely, Problem Transformation (PT) and Algorithm Adaptation (AA) methods. This categorization is extended to include a third category of methods by Gjorgji Madjarov et al [18] called Ensemble methods (EN). Several review articles are available in the literature that describe various methods available for multi-label classification [7,8,15,17,18]. As adapted from [18], an overview of multi-label methods available in the literature is given in Fig. 1.

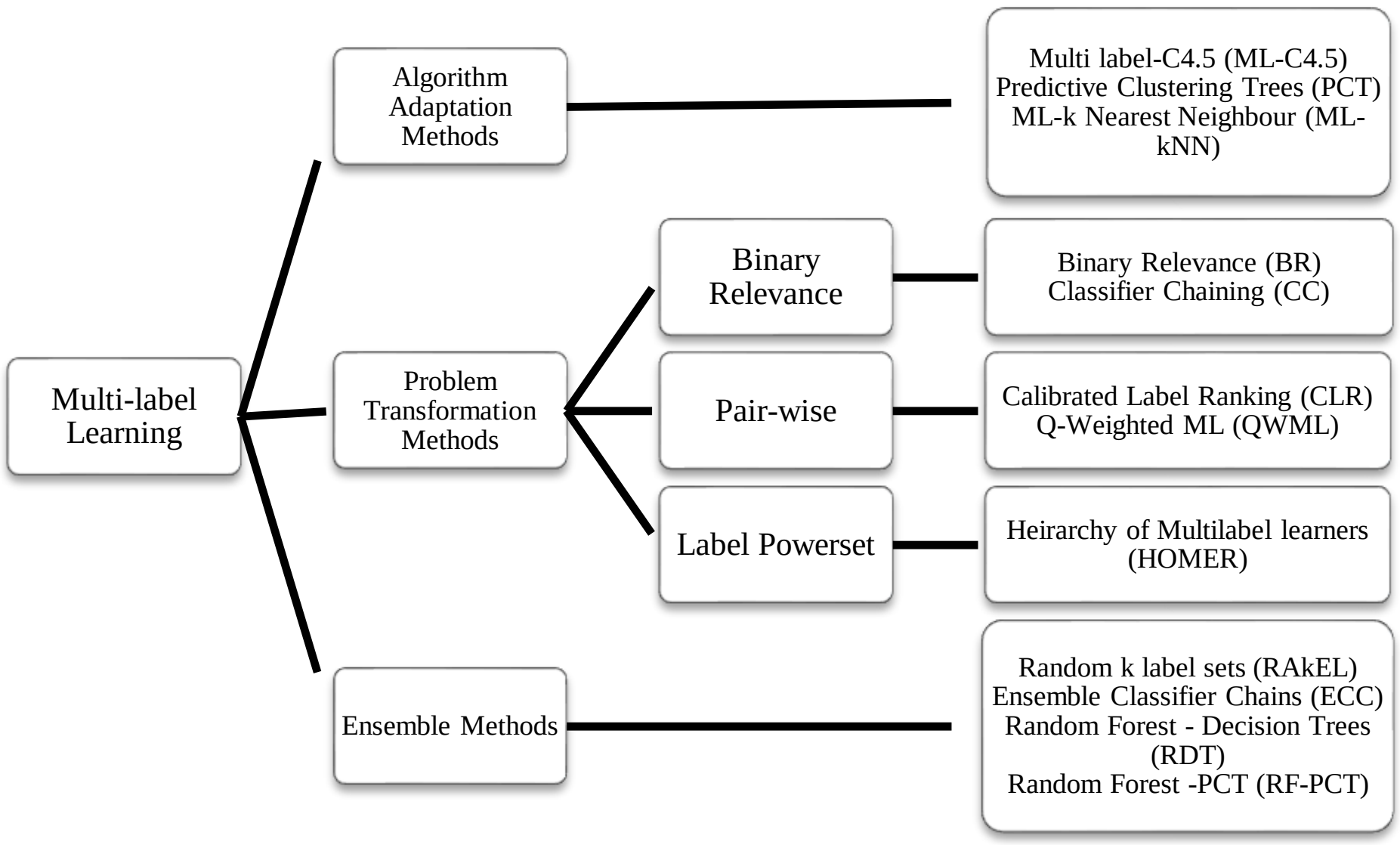

**Fig. 1.** Classification of multi-label methods

Based on the machine learning algorithm used, the multi-label techniques can be categorized as shown in Fig. 2, adapted from [18]. This paper proposes a high speed multi-label learning technique based on ELM, which outperforms all the existing techniques based on speed and performance.

## 3 Proposed Approach

The extreme learning machine is a learning technique that operates on a single-layer feedforward neural network. The key advantage of the ELM over the traditional back-propagation (BP) neural network is that it has the smallest number of parameters to be adjusted and it can be trained with very high speed. The traditional BP network needs to be initialized and several parameters tuned and improper selection of which can result in local optima. On the other hand, in ELM, the initial weights and the hidden layer bias can be selected at random and the network can be trained for the output weights in order to perform the classification [19-22]. The key steps in extending the ELM to multi-label problems is in the pre-processing and post-processing of data. In multi-label problems, each input sample may belong to one or more samples. The number of labels an input sample belongs to is not previously known. Therefore, both the number of labels and the target labels are to be identified for the test input samples and also the degree of multi-labelness varies among different datasets. This results in increased complexity of the multi-label problem resulting in much longer training and testing time of the multi-label classification technique. The proposed algorithm exploits the inherent high speed nature of the ELM resulting in both high speed and superior performance compared with the existing multi-label classification techniques.

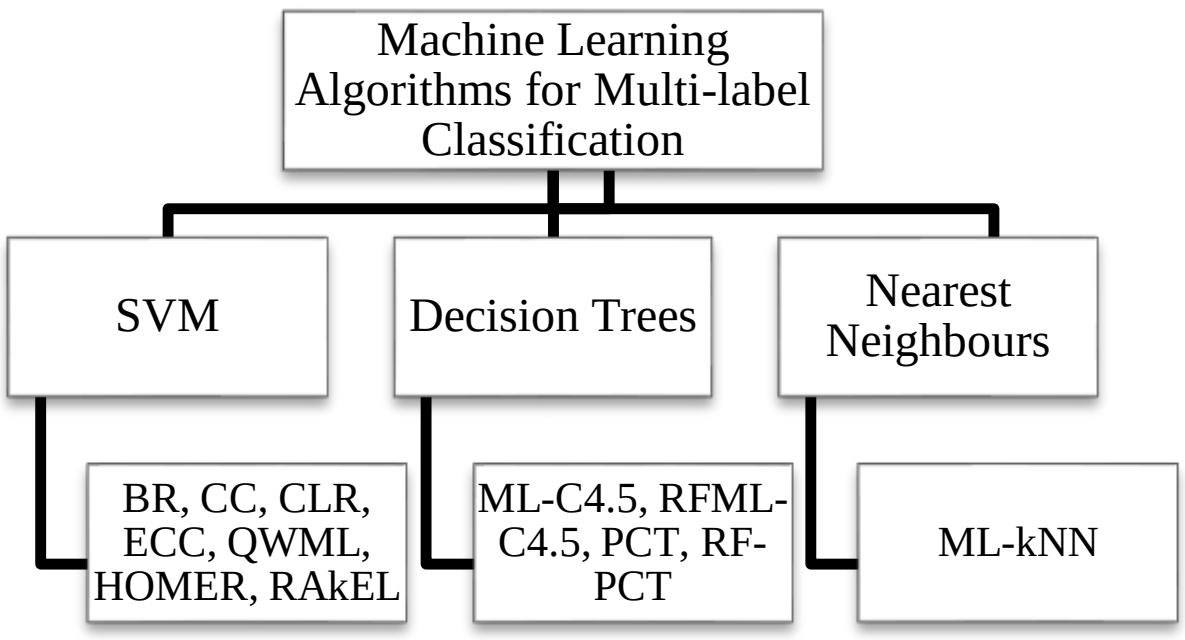

**Fig. 2.** Machine learning algorithms for multi-label problems

Consider N training samples of the form $\{(x_i,y_i)\}$ where $x_i$ in the input denoted as $x_i = [x_{i1},x_{i2},\ldots,x_{in}]^T \in R^n$ and $y_i$ is the target label set, $y_i = [y_{i1},y_{i2},\ldots y_{im}]^T$. As opposed to traditional single-label case, the target label is not a single label but is a subset of labels from the label space given as $Y \subseteq L$, $L = \{\zeta_1, \zeta_2, \ldots, \zeta_M\}$. Let $\bar{N}$ be the number of hidden layer neurons, the output 'o' of the SLFN is given by

$$\sum_{i=1}^{\bar{N}} \beta_i g_i(x_j) = \sum_{i=1}^{\bar{N}} \beta_i g(w_i \,.\, x_j + b_i) = o_j \quad (2)$$

where, $\beta_i = [\beta_{i1},\beta_{i2},\ldots\beta_{im}]^T$ is the output weight, g(x) is the activation function, $w_i = [w_{i1},w_{i2},\ldots w_{in}]T$ is the input weight and bi is the hidden layer bias.

For the ELM, the input weights $w_i$ and the hidden layer bias bi are randomly assigned. Therefore, the network must be trained for βi such that the output of the network is equal to the target class so that the error difference between the actual output and the predicted output is 0.

$$\sum_{j=1}^{\bar{N}} \left\| o_j - y_j \right\| = 0 \quad (3)$$

Thus, the ELM classifier output can be as follows:

$$\sum_{i=1}^{\bar{N}} \beta_i g(w_i \,.\, x_j + b_i) = y_j \quad (4)$$

The above equation can be written in following matrix form:

$$H\beta = Y \quad (5)$$

The output weights of the ELM network can be estimated using the equation

$$\beta = H^{+}Y \quad (6a)$$

where $H^+$ is the Moore-Penrose inverse of the hidden layer output matrix H and it can be calculated as follows:

$$H^+ = (H^T H)^{-1} H^T \tag{6b}$$

The theory and mathematics behind the ELM have been extensively discussed in [23-25] and hence are not re-stated here. The steps involved in multi-label ELM classifier are given below.

*Initialization of Parameters.* Fundamental parameters such as the number of hidden layer neurons and the activation function are initialized.

*Processing of Inputs.* In the multi-label case, each input sample can be associated with more than one class labels. Hence, each of the input samples will have the associated output label as a m-tuple with 0 or 1 representing the belongingness to each of the labels in the label space L. The label set denoting the belongingness for each of the labels is converted from unipolar representation to bipolar representation.

*ELM Training.* The processed input is then supplied to the basic batch learning ELM. Let H be the hidden layer output matrix, β be the output weights and Y be the target label, the ELM can be represented in a compact form as $H\beta = Y$ where $Y \subseteq L$, $L = \{\zeta_1, \zeta_2, \ldots, \zeta_M\}$. In the training phase, the input weights and the hidden layer bias are randomly assigned and the output weights β are estimated as $\beta = H^+Y$, where $H^+ = (H^T H)^{-1}H^T$ gives the Moore-Penrose generalized inverse of the hidden layer output matrix.

*ELM Testing.* In the testing phase, the test data sample is evaluated using the values of β obtained during the training phase. The network then predicts the target output using the equation $Y = H\beta$. The predicted output Y obtained is a set of real numbers of dimension equal to the number of labels.

*Post-processing and Multi-label Identification.* The key challenge in multi-label classification is that the input sample may belong to one or more than one of the target labels. The number of labels that the sample corresponds to is completely unknown. Hence, a thresholding-based label association is proposed. The L dimensioned raw-predicted output is compared with a threshold value. The index values of the predicted output Y which are greater than the threshold fixed represents the belongingness of the input sample to the corresponding class.

Setting the threshold value is of critical importance. Threshold setting has to be made in such a way that it maximizes the difference between the values of the label the data belongs to and the labels the data does not. The distribution of the raw output values is categorized into a range of values that represent the belongingness of the label and the range of values that represent the non-belongingness of the label to a particular sample. From the distribution, a particular value is chosen that maximizes the separation between the two categories of the labels. It is to be highlighted that there are no ELM-based multi-label classifier in the literature thus far. The proposed method is the first to adapt the ELM for multi-label problems and make extensive experimentation and results comparison and analysis with the state-of-the-arts multi-label classification techniques.

## 4 Experimentation

This section describes the different multi-label dataset metrics and gives the experimental design used to evaluate the proposed method. Multi-label datasets have a unique property called the degree of multi-labelness. The number of labels, the number of samples having multiple labels, the average number of labels corresponding to a particular sample varies among different datasets. Two dataset metrics are available in the literature to quantitatively measure the multi-labelness of a dataset. They are Label Cardinality (LC) and Label Density (LD). Consider there are N training samples and the dataset is of the form $\{(x_i,y_i)\}$ where $x_i$ in the input data and $y_i$ is the target label set. The target label set is a subset of labels from the label space with M elements given as $Y \subseteq L$, $L = \{\zeta_1, \zeta_2 \ldots \zeta_M\}$.

***Definition 4.1*** [17] *Label Cardinality of the dataset is the average number of labels of the examples in the dataset.*

$$Label - Cardinality = \frac{1}{N}\sum_{i=1}^{N}|Y_i| \tag{7}$$

Label Cardinality signifies the average number of labels present in the dataset.

***Definition 4.2*** [17] *Label Density of the dataset is the average number of labels of the examples in the dataset divided by* $|L|$.

$$Label - Density = \frac{1}{N}\sum_{i=1}^{N}\frac{|Y_i|}{|L|} \tag{8}$$

Label density takes into consideration the number of labels present in the dataset. The properties of two datasets have same label cardinality, but different label density can vary significantly and may result in different behavior of the training algorithm [14]. The influence of label density and label cardinality on multi-label learning is analyzed by Flavia et al in 2013 [26]. The proposed method is experimented with six benchmark datasets comprising of different application areas and its results are compared with 9 existing state-of-the-art methods. The datasets are chosen in such a way that they exhibit diverse nature of characteristics and the wide range of label density and label cardinality. The datasets are obtained from KEEL multi-label dataset repository and the specifications of the dataset are given in Table 1. The details of state-of-the-arts multi-label techniques used for result comparison are given in Table 2.

**Table 1.** Dataset specifications

| Dataset | Domain | No. of Features | No. of Samples | No. of Labels | LC | LD |
|---|---|---|---|---|---|---|
| **Emotion** | Multimedia | 72 | 593 | 6 | 1.87 | 0.312 |
| **Yeast** | Biology | 103 | 2417 | 14 | 4.24 | 0.303 |
| **Scene** | Multimedia | 294 | 2407 | 6 | 1.07 | 0.178 |
| **Corel5k** | Multimedia | 499 | 5000 | 374 | 3.52 | 0.009 |
| **Enron** | Text | 1001 | 1702 | 53 | 3.38 | 0.064 |
| **Medical** | Text | 1449 | 978 | 45 | 1.25 | 0.027 |

**Table 2.** Methods used for comparison

| Method Name | Method Category | Machine Learning Category |
|---|---|---|
| **Classifier Chain (CC)** | PT | SVM |
| **QWeighted approach for Multi-label Learning (QWML)** | PT | SVM |
| **Hierarchy Of Multi-label ClassifiERs (HOMER)** | PT | SVM |
| **Multi-Label C4.5 (ML-C4.5)** | AA | Decision Trees |
| **Predictive Clustering Trees (PCT)** | AA | Decision Trees |
| **Multi-Label k-Nearest Neighbors (ML-kNN)** | AA | Nearest Neighbors |
| **Ensemble of Classifier Chains (ECC)** | EN | SVM |
| **Random Forest Predictive Clustering Trees (RF-PCT)** | EN | Decision Trees |
| **Random Forest of ML-C4.5 (RFML-C4.5)** | EN | Decision Trees |

# 5 Results and Discussions

This section discusses the results obtained by the proposed method and compares it with the existing methods. The results obtained from the proposed method are evaluated for consistency, performance and speed.

## 5.1 Consistency

Consistency is a key feature that is essential for any new technique proposed. The proposed algorithm should provide consistent results with minimal variance. Being an ELM based algorithm, since the initial weights are assigned in random, it is critical to evaluate the consistency of the proposed technique. The unique feature of multi-label classification is the possibility of partial correctness of the classifier, i.e. one or more of the multiple labels to which the sample instance belongs and/or the number of labels the sample instance belongs can be identified partially correctly. Therefore, calculating the error rate for multi-label problems is not same as that of traditional binary or multi-class problems. In order to quantitatively measure the correctness of the classifier, the hamming loss performance metric is used. To evaluate the consistency of the proposed method, a 5 fold and a 10 fold cross validation of hamming loss metric is evaluated for each of the six datasets and is tabulated.

**Table 3.** Consistency table – cross validation

| Dataset | Hamming Loss - 5-fcv | Hamming Loss - 10-fcv |
|---|---|---|
| **Emotion** | 0.2492(±0.0058) | 0.2509(±0.0050) |
| **Yeast** | 0.1906(±0.0025) | 0.1911(±0.0031) |
| **Scene** | 0.0854(±0.0029) | 0.0851(±0.0033) |
| **Corel5k** | 0.0086(±0.0005) | 0.0090(±0.0006) |
| **Enron** | 0.0474(±0.0022) | 0.0472(±0.0015) |
| **Medical** | 0.0108(±0.0008) | 0.0109(±0.0009) |

From the table 3, it can be seen that the proposed technique is consistent in its performance over repeated executions and cross validations thus demonstrating the consistency of the technique.

## 5.2 Performance Metrics

As foreshadowed, the unique feature of multi-label classification is the possibility of partial correctness of the classifier. Therefore, a set of quantitative performance evaluation metrics is used to validate the performance of the multi-label classifier. The performance metrics are hamming loss, accuracy, precision, recall and F1-measure. A comparison of performance metrics such as hamming loss, precision, recall, accuracy and F1 measure of the proposed technique is shown in Tables 4-8. The performance of state-of-the-art techniques is adapted from [18]. From the tables, it is clear that the proposed method works uniformly well on all datasets. The proposed method outperforms all the existing methods in most cases and remains one of the top classification techniques in other cases.

**Table 4.** Hamming loss comparison

| Dataset | CC | QWML | HOMER | ML-C4.5 | PCT | ML-kNN | ECC | RFML-C4.5 | RF-PCT | ELM |
|---|---|---|---|---|---|---|---|---|---|---|
| **Emotion** | 0.256 | 0.254 | 0.361 | 0.247 | 0.267 | 0.294 | 0.281 | 0.198 | 0.189 | **0.251** |
| **Yeast** | 0.193 | 0.191 | 0.207 | 0.234 | 0.219 | 0.198 | 0.207 | 0.205 | 0.197 | **0.191** |
| **Scene** | 0.082 | 0.081 | 0.082 | 0.141 | 0.129 | 0.099 | 0.085 | 0.116 | 0.094 | **0.085** |
| **Corel 5k** | 0.017 | 0.012 | 0.012 | 0.01 | 0.009 | 0.009 | 0.009 | 0.009 | 0.009 | **0.009** |
| **Enron** | 0.064 | 0.048 | 0.051 | 0.053 | 0.058 | 0.051 | 0.049 | 0.047 | 0.046 | **0.047** |
| **Medical** | 0.077 | 0.012 | 0.012 | 0.013 | 0.023 | 0.017 | 0.014 | 0.022 | 0.014 | **0.011** |

**Table 5.** Accuracy comparison

| Dataset | CC | QWML | HOMER | ML-C4.5 | PCT | ML-kNN | ECC | RFML-C4.5 | RF-PCT | ELM |
|---|---|---|---|---|---|---|---|---|---|---|
| **Emotion** | 0.356 | 0.373 | 0.471 | 0.536 | 0.448 | 0.319 | 0.432 | 0.488 | 0.519 | **0.412** |
| **Yeast** | 0.527 | 0.523 | 0.559 | 0.48 | 0.44 | 0.492 | 0.546 | 0.453 | 0.478 | **0.514** |
| **Scene** | 0.723 | 0.683 | 0.717 | 0.569 | 0.538 | 0.629 | 0.735 | 0.388 | 0.541 | **0.676** |
| **Corel 5k** | 0.03 | 0.195 | 0.179 | 0.002 | 0 | 0.014 | 0.001 | 0.005 | 0.009 | **0.044** |
| **Enron** | 0.334 | 0.388 | 0.478 | 0.418 | 0.196 | 0.319 | 0.462 | 0.374 | 0.416 | **0.418** |
| **Medical** | 0.211 | 0.658 | 0.713 | 0.73 | 0.228 | 0.528 | 0.611 | 0.25 | 0.591 | **0.715** |

**Table 6.** Precision comparison

| Dataset | CC | QWML | HOMER | ML-C4.5 | PCT | ML-kNN | ECC | RFML-C4.5 | RF-PCT | ELM |
|---|---|---|---|---|---|---|---|---|---|---|
| **Emotion** | 0.551 | 0.548 | 0.509 | 0.606 | 0.577 | 0.502 | 0.58 | 0.625 | 0.644 | **0.548** |
| **Yeast** | 0.727 | 0.718 | 0.663 | 0.62 | 0.705 | 0.732 | 0.667 | 0.738 | 0.744 | **0.718** |
| **Scene** | 0.758 | 0.711 | 0.746 | 0.592 | 0.565 | 0.661 | 0.77 | 0.403 | 0.565 | **0.685** |
| **Corel 5k** | 0.042 | 0.326 | 0.317 | 0.005 | 0 | 0.035 | 0.002 | 0.018 | 0.03 | **0.144** |
| **Enron** | 0.464 | 0.624 | 0.616 | 0.623 | 0.415 | 0.587 | 0.652 | 0.69 | 0.709 | **0.668** |
| **Medical** | 0.217 | 0.697 | 0.762 | 0.797 | 0.285 | 0.575 | 0.662 | 0.284 | 0.635 | **0.774** |

**Table 7.** Recall comparison

| Dataset | CC | QWML | HOMER | ML-C4.5 | PCT | ML-kNN | ECC | RFML-C4.5 | RF-PCT | ELM |
|---|---|---|---|---|---|---|---|---|---|---|
| **Emotion** | 0.397 | 0.429 | 0.775 | 0.703 | 0.534 | 0.377 | 0.533 | 0.545 | 0.582 | **0.491** |
| **Yeast** | 0.6 | 0.6 | 0.714 | 0.608 | 0.49 | 0.549 | 0.673 | 0.491 | 0.523 | **0.608** |
| **Scene** | 0.726 | 0.709 | 0.744 | 0.582 | 0.539 | 0.655 | 0.771 | 0.388 | 0.541 | **0.709** |
| **Corel5k** | 0.056 | 0.264 | 0.25 | 0.002 | 0 | 0.014 | 0.001 | 0.005 | 0.009 | **0.043** |
| **Enron** | 0.507 | 0.453 | 0.61 | 0.487 | 0.229 | 0.358 | 0.56 | 0.398 | 0.452 | **0.508** |
| **Medical** | 0.754 | 0.801 | 0.76 | 0.74 | 0.227 | 0.547 | 0.642 | 0.251 | 0.599 | **0.744** |

**Table 8.** F1 measure comparison

| Dataset | CC | QWML | HOMER | ML-C4.5 | PCT | ML-kNN | ECC | RFML-C4.5 | RF-PCT | ELM |
|---|---|---|---|---|---|---|---|---|---|---|
| **Emotion** | 0.461 | 0.481 | 0.614 | 0.651 | 0.554 | 0.431 | 0.556 | 0.583 | 0.611 | **0.518** |
| **Yeast** | 0.657 | 0.654 | 0.687 | 0.614 | 0.578 | 0.628 | 0.67 | 0.589 | 0.614 | **0.658** |
| **Scene** | 0.742 | 0.71 | 0.745 | 0.587 | 0.551 | 0.658 | 0.771 | 0.395 | 0.553 | **0.697** |
| **Corel 5k** | 0.048 | 0.292 | 0.28 | 0.003 | 0 | 0.021 | 0.001 | 0.008 | 0.014 | **0.033** |
| **Enron** | 0.484 | 0.525 | 0.613 | 0.546 | 0.295 | 0.445 | 0.602 | 0.505 | 0.552 | **0.577** |
| **Medical** | 0.337 | 0.745 | 0.761 | 0.768 | 0.253 | 0.56 | 0.652 | 0.267 | 0.616 | **0.759** |

## 5.3 Speed

The performance of the proposed method in terms of execution speed is evaluated by comparing the training time and the testing time of the algorithm used. The proposed method is applied to 6 datasets of different domains with a wide range of label density and label cardinality and the training time and the testing time are compared with other state-of-the-art techniques. The comparison table of training time and testing time is given in Table 9 and Table 10 respectively.

**Table 9.** Comparison of training time (in seconds)

| Dataset | CC | QWML | HOMER | ML-C4.5 | PCT | ML-kNN | ECC | RFML-C4.5 | RF-PCT | ELM |
|---|---|---|---|---|---|---|---|---|---|---|
| **Emotion** | 6 | 10 | 4 | 0.3 | 0.1 | 0.4 | 4.9 | 1.2 | 2.9 | **0.04** |
| **Yeast** | 206 | 672 | 101 | 14 | 1.5 | 8.2 | 497 | 19 | 25 | **0.2** |
| **Scene** | 99 | 195 | 68 | 8 | 2 | 14 | 319 | 10 | 23 | **0.12** |
| **Corel5k** | 1225 | 2388 | 771 | 369 | 30 | 389 | 10073 | 385 | 902 | **0.6** |
| **Enron** | 440 | 971 | 158 | 15 | 1.1 | 6 | 1467 | 25 | 47 | **0.26** |
| **Medical** | 28 | 40 | 16 | 3 | 0.6 | 1 | 103 | 7 | 27 | **0.11** |

**Table 10.** Comparison of testing time (in seconds)

| Dataset | CC | QWML | HOMER | ML-C4.5 | PCT | ML-kNN | ECC | RFML-C4.5 | RF-PCT | ELM |
|---|---|---|---|---|---|---|---|---|---|---|
| **Emotion** | 1 | 2 | 1 | 0 | 0 | 0.4 | 6.6 | 0.1 | 0.3 | **0** |
| **Yeast** | 25 | 64 | 17 | 0.1 | 0 | 5 | 158 | 0.5 | 0.2 | **0** |
| **Scene** | 25 | 40 | 21 | 1 | 0 | 14 | 168 | 2 | 1 | **0** |
| **Corel5k** | 31 | 119 | 14 | 1 | 1 | 45 | 2077 | 1.8 | 2.5 | **0.06** |
| **Enron** | 53 | 174 | 22 | 0.2 | 0 | 3 | 696 | 1 | 1 | **0** |
| **Medical** | 6 | 25 | 1.5 | 0.1 | 0 | 0.2 | 46 | 0.5 | 0.5 | **0** |

In summary, the proposed method outperforms all existing multi-label learning techniques in terms of training and testing time by several orders of magnitude. From the results, it can be seen that the proposed method is the fastest multi-label classifier when compared to the current state-of-the-arts techniques. The speed of the proposed classifier is many-fold greater than existing methods. Also, from the comparison results of other performance metrics such as hamming loss, accuracy, precision, recall and F1-measure, it can be seen that the proposed method remains one of the top positions in each case. Also, the F1-measure of the proposed approach outperforms the most recent method which uses canonical correlation analysis (CCA) with ELM for multi-label problems [27] in most cases. The key advantage of the proposed method is that it surpasses all existing state-of-the-arts methods in terms of speed and simultaneously while remaining one of the top learning techniques in terms of other 5 performance metrics.

## 6 Conclusion

The proposed high speed multi-label classifier executes with both fast speed and high accuracy. It is to be highlighted that there are no extreme-learning-machine-based multi-label classifiers existing in the literature thus far. The proposed method is applied to 6 benchmark datasets of different domains and a wide range of label density and label cardinality. The results are compared with 9 state-of-the-arts multi-label classifiers. It can be seen from the results that the proposed method surpasses all state-of-the-arts methods in terms of speed and remain one of the top techniques in terms of other performance metrics. Thus, the proposed ELM-based multi-label classifier can be a better alternative for a wide range of multi-label classification techniques in order to achieve greater accuracy and very high speed.

**Acknowledgements**

This work is supported by the National Natural Science Foundation of P. R. China (under Grants 51009017 and 51379002), Applied Basic Research Funds from Ministry of Transport of P. R. China (under Grant 2012-329-225-060), China Postdoctoral Science Foundation (under Grant 2012M520629), Program for Liaoning Excellent Talents in University (under Grant LJQ2013055). The second author would like to thank Nanyang Technological University for supporting this work by providing NTU RSS.

## References

1. T. Gonclaves, P. Quaresma, "A Preliminary approach to the multi-label classification problem of Portuguese juridical documents", Progress in Artificial Intelligence, Springer Berlin Heidelberg, pp. 435-444, 2003.
2. T. Joachims, "Text categorization with support vector machines: Learning with many relevant features", Nedellec C, Rouveirol C (Ed.), ECML, LNCS, vol. 1938, pp. 137-142, Springer, Heidelberg, 1998.
3. X. Luo, A. N. Zincir Heywood, "Evaluation of Two Systems on Multi-class Multi-label document classification", Hacid M.S., Murray N.V., Ras Z.W., Tsumoto S (Ed.), ISMIS 2005, LNCS (LNAI), vol. 3488, pp. 161-169, Springer, Heidelberg, 2005.
4. D. Tikk, G Biro, "Experiments with multi-label text classifier on the Reuters collection", Proceedings of the International Conference on Computational Cybernetics (ICCC 2003), Hungary, pp. 33-38, 2003.
5. K. Yu, S. Yu, V. Tresp, "Multi-label informed latent semantic indexing", Proceedings of the 28th annual international ACM SIGIR conference on Research and Development in information retrieval, pp. 258-265, 2005.
6. A. Karalic, V. Pirnat, "Significance level based multiple tree classification", Informatica, vol. 15(5), 12 pages, 1991.
7. G. Tsoumakas, I. Katakis, I. Vlahavas, "Mining Multi-label Data", Data Mining and Knowledge Discovery Handbook, O. Maimon, L. Rokach (Ed.), Springer, 2nd Edition, 2010.

8. A. C. de Carvalho, A. A. Freitas, “A Tutorial on Multi-label Classification Techniques”, Foundations of Computational Intelligence, vol. 5, pp. 177-195, 2009.
9. A. Elisseeff, J. Weston, “A kernel method for multi-labelled classification”, Neural Information Processing Systems, NIPS, vol. 14, 2001.
10. M. L. Zhang, Z. H. Zhou, “A k-nearest neighbour based algorithm for multi-label classification”, Proceedings of the 1st IEEE international Conference on Granular Computing, Beijing, China, pp. 718-721, 2005.
11. M. Boutell, X. Shen, J. Luo, C. Brouwn, “Multi-label semantic scene classification”, Technical Report, Department of Computer Science University of Rochester, USA, 2003.
12. X. Shen, M. Boutell, J. Luo, C. Brown, “Multi-label machine learning and its application to semantic scene classification. Storage and Retrieval Methods and Applications for Multimedia”, Yeung MM, Lienhart RW, Li CS (Ed.), Proceedings of the SPIE, vol. 5307, pp. 188-199, 2003.
13. B. Zhu, C. K. Poon, “Efficient Approximation Algorithms for Multi-label Map Labelling”, Algorithms and Computation, pp. 143-152, Springer Heidelberg, 1999.
14. M. L. Zhang, Z. H. Zhou, “ML-kNN: A lazy learning approach to multi-label learning”, Pattern Recognition, vol. 40(7), pp. 2038-2048, 2007.
15. M. S. Sorower, "A literature survey on algorithms for multi-label learning." Oregon State University, Corvallis, 2010.
16. A. Elisseeff, J. Weston J, “Kernel methods for multi-labelled classification and categorical regression problems”, Technical Report, BIOwulf Technologies, 2001.
17. G. Tsoumakas, I. Katakis, “Multi-label Classification: An Overview”, International Journal of Data Warehousing and Mining, vol. 3(3), pp. 1-13, 2007.
18. G. Madjarov, D. Kocev, D. Gjorgjevikj, S. Dzeroski, “An extensive experimental comparison of methods for multi-label learning”, Pattern Recognition, vol. 45, pp. 3084-3104, 2012.
19. N. Wang, J. C. Sun, M. J. Er and Y. C. Liu, “A novel extreme learning control framework of unmanned surface vehicles”, IEEE Transactions on Cybernetics, Accepted for Publication, 2015.
20. N. Wang, M. J. Er, M. Han, “Generalized single-hidden layer feedforward networks for regression problems”, IEEE Transactions on Neural Networks and Learning Systems, vol. 26, no. 6, pp. 1161-1176, 2015.
21. N. Wang, M. J. Er, M. Han, “Parsimonious extreme learning machine using recursive orthogonal least squares”, IEEE Transactions on Neural Networks and Learning Systems, vol. 25, no. 10, pp. 1828-1841, 2014.
22. N. Wang, M. Han, N. Dong, M. J. Er, “Constructive multi-output extreme learning machine with application to large tanker motion dynamics identification”, Neurocomputing, vol. 128, pp. 59-72, 2014.
23. G. B. Huang, D. H. Wang, Y. Lan., "Extreme learning machines: a survey," International Journal of Machine Learning and Cybernetics, vol. 2, pp. 107-122, 06/01, 2011.
24. G. B. Huang, Q. Y. Zhu, C. K. Siew, "Extreme learning machine: Theory and applications," Neurocomputing, vol. 70, pp. 489-501, 12// 2006.
25. S. Ding, H. Zhao, Y. Zhang, X. Xu, R. Nie, “Extreme learning machine: algorithm, theory and applications”, Artificial Intelligence Review, 1-13, 2013.
26. F. C. Bernardini, R. B. da Silva, E. M. Meza, das Ostras–RJ–Brazil, R. “Analyzing the Influence of Cardinality and Density Characteristics on Multi-Label Learning”, 2009.
27. Y. Kongsorot, P. Horata, “Multi-label Classification with Extreme Learning Machine”, International Conference on Knowledge and Smart Technology, pp. 81-86.